\documentclass[a4paper]{article}

\usepackage{microtype}
\usepackage{graphicx}
\usepackage{subfigure}
\usepackage{booktabs} 
\usepackage{natbib}
\usepackage{hyperref}

\usepackage{amsmath}
\usepackage{amssymb}
\usepackage{mathtools}
\usepackage{amsthm}

\usepackage[capitalize,noabbrev]{cleveref}

\theoremstyle{plain}

\theoremstyle{definition}
\newtheorem{definition}{Definition}

\theoremstyle{remark}
\newtheorem{remark}{Remark}[section]

\usepackage[a4paper, total={5.5in, 8in}]{geometry}

\title{Preparing for Black Swans \includegraphics[height=1em]{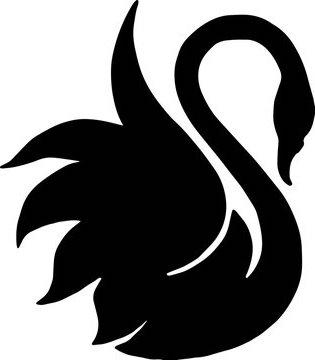}: \\
The Antifragility Imperative for Machine Learning}

\author{Ming Jin\thanks{This work was supported in part by the NSF Safe Learning-Enabled Systems Program (NSF \# 2331775).}
}
\date{} 

\begin{document}
\maketitle

\begin{abstract}
Operating safely and reliably despite continual distribution shifts is vital for high-stakes machine learning applications. This paper builds upon the transformative concept of ``antifragility'' introduced by \citep{taleb2014antifragile} as a constructive design paradigm to not just withstand but benefit from volatility. We formally define antifragility in the context of online decision making as dynamic regret's strictly concave response to environmental variability, revealing limitations of current approaches focused on resisting rather than benefiting from nonstationarity. Our contribution lies in proposing potential computational pathways for engineering antifragility, grounding the concept in online learning theory and drawing connections to recent advancements in areas such as meta-learning, safe exploration, continual learning, multi-objective/quality-diversity optimization, and foundation models. By identifying promising mechanisms and future research directions, we aim to put antifragility on a rigorous theoretical foundation in machine learning. We further emphasize the need for clear guidelines, risk assessment frameworks, and interdisciplinary collaboration to ensure responsible application.

\end{abstract}

\section{Introduction}

As we integrate machine learning (ML) into mission-critical systems in healthcare, finance, transportation, and social-scale infrastructure like power grids, a vital question arises about ensuring safety, security, and reliability despite myriad stressors
\citep{hendrycks2021unsolved}. These manifest as natural or adversarial perturbations, aleatoric/epistemic uncertainties, distribution shifts (including domain shift, concept drift, nonstionarity, and out-of-distribution events).

The prevailing paradigm in ML pursues robustness---shielding systems against these stressors. However, as S. Sagan cautions, ``Things that have never happened before happen all the time.'' Novel outliers routinely breach reactive defenses, raising concerns about their limitations against rare but impactful ``black swan'' events \citep{taleb2010black}. \citep{nair2022fundamentals} present statistical arguments about why such long-tailed phenomena prove unexpectedly ubiquitous. 

\begin{quote}
    Rather than merely withstand stressors, can systems be designed to thrive because of them?
\end{quote}

This inverted perspective of embracing rather than barricading against disorder constitutes the radical notion of ``antifragility.'' As originally defined by \citep{taleb2014antifragile}: 
\begin{quote}
    ``Some things benefit from shocks; they thrive and grow when exposed to volatility, randomness, disorder, and stressors and love adventure, risk, and uncertainty. Yet, in spite of the ubiquity of the phenomenon, there is no word for the exact opposite of fragile. Let us call it antifragile. Antifragility is beyond resilience or robustness. The resilient resists shocks and stays the same; the antifragile gets better. ''
\end{quote}

Despite inspiring examples across various domains, formal frameworks in ML leveraging antifragility remain scarce. This manuscript proposes a rigorous definition of antifragility based on the online decision-making (ODM) framework and identifies potential computational pathways to realize this vision.

\subsection{Perspective}

We see antifragility as both an attainable goal and a constructive design solution for ML systems.

Antifragile systems embody an open-world perspective, actively leveraging continual stressors such as distributional shifts to systematically transform them into opportunities for enhancement. These systems are capable not just of adaptation, but of \emph{rapid} adaptation with improvisation when faced with unforeseen disruptions. They exhibit \emph{unusual} robustness that goes beyond typical stressors to confront potentially unforeseeable, high-impact challenges. This is achieved through proactive preparation for such (unknown) unknowns by exposure to and learning from a \emph{diverse} set of prior disturbances. By investigating computational mechanisms that enable systems to benefit from stressors and grounding the concept in rigorous theoretical frameworks, we can collectively work towards computational antifragility for ML systems that thrive amidst volatility. 

\subsection{Rejoinders}

Antifragility, while a transformative concept, understandably faces skepticism---reliability and security demand assurance while random tinkering seems reckless. Volatility may productively ``vaccinate,'' but its dangers cannot be casually dismissed, especially in domains where errors carry irreversible consequences.\footnote{In this note, ``volatility'' broadly include 2nd or higher order moments of a random variable, \emph{variability} within time series, and distributional shifts (nonstationarity).}

We do not argue for fully embracing disorder or eliminating all conservative controls, which could decrease safety and oversight. Instead, we advocate intelligent, measured openness through safe environment interaction, selectively harnessing volatility’s benefits under safety constraints. This purposeful experimentation permits recoverable failures that can be safely remediated (see Sec. \ref{sec: safe} for some methods). In critical domains like healthcare and cyber-physical infrastructures, methods that simulate and small-scale test disturbances provide safe avenues. Further, natural perturbations may already provide ample opportunities for learning (even more than engineered ones), reducing risks of intentional volatility. Hence, antifragility should not be misconstrued as compromising safety; rather, it enlarges its boundaries through interactively testing and evolving risk tolerance. Appendix~\ref{app:safe} discusses guidelines and risk assessment frameworks for responsible and ethically-informed application of antifragility principles.

Antifragility does not oppose but complements robustness, benefiting as systems dynamically navigate and strengthen through smaller challenges. Robustness, whether through simple redundancy or safe controls, provides a safe net enabling guided volatility. However, antifragility extends beyond immediate reactions to stressors; it operates on longer timespans under larger magnitudes, learning and evolving especially when robustness falters against surprises. Hence, it develops a  viable path towards \emph{unusually} robust systems by dispersing risks across smaller, manageable events to preempt catastrophic failures during rare but inevitable black swan events.  Robustness sustains today's operations; antifragility invests in tomorrow's upside. App. \ref{app:related} further discusses connections to related paradigms such as robustness, lifelong learning, and cognitive science.

\subsection{Evidence}

The viability of antifragility can be observed in various natural and engineered systems. While the internal mechanisms enabling antifragility can be complex and may result from long-term evolution, we can identify antifragile properties by examining a system's response to stressors over time. Several examples, spanning biological, economic, social, and technological domains, are briefly described below and further detailed in App. \ref{app:examples}.

\subsection{Overview}

The conceptual scope of this manuscript is framed through online learning, an area with long history and contemporary relevance at the core of ODM that has seen much cross-pollination with related fields such as optimization and reinforcement learning (RL). We believe this provides a solid foundation for establishing antifragility given its matching characteristic of nonstationarity. Specifically, we leverage the well-defined dynamic regret to provide a rigorous definition of antifragility (Sec. \ref{sec:definition}) and characterize some fundamental impediments and open problems (Sec. \ref{sec:lowerbound}). Through connections to RL and its variants, representing typical interactive settings, as well as areas like meta-learning, continual learning, and foundation models among others, we aim to spark constructive discussions (Sec.~\ref{sec:discussion}), rally momentum to address gaps (Sec. \ref{sec:future}), and spur work towards computational antifragility.

\section{Preliminaries}
\label{sec:prelim}
Online learning tackles sequential decisions based on live data streams, gaining increasing relevance amid real-time needs \citep{hoi2021online}. Foundational introductions are provided in \citep{shalev2012online,hazan2016introduction,orabona2019modern}.
The setup involves an action set $\mathcal{A}$, a time horizon $T$, and a sequence of cost functions $\ell_1, \ldots, \ell_T$. For any comparator $u \in \mathcal{A}$, the static regret is:
\begin{equation}
    \label{def:static-regret}
    R_T(u) \triangleq \sum_{t=1}^T \ell_t(a_t) - \sum_{t=1}^T \ell_t(u),
\end{equation}
where $a_t$ is the algorithm's output at time $t$. Unlike static regret, the dynamic regret is defined against a sequence of comparators $u_{1:T} = (u_1, \ldots, u_T)$ as:
\begin{equation}
    \label{def:dynamic-regret}
    R_T(u_{1:T}) \triangleq \sum_{t=1}^T \ell_t(a_t) - \sum_{t=1}^T \ell_t(u_t).
\end{equation}
Dynamic regret is particularly relevant in nonstationary environments. Two common measures for modeling the environment's stationarity include: 1) Path-Length: Introduced by \citep{zinkevich2003online}, the path-length of the sequence $u_{1:T}$ is defined as: $ V^u_T(u_{1:T}) \triangleq \sum_{t=2}^T \rho( u_t, u_{t-1}),$ where $\rho$ is a metric that measures the distance between $u_t$ and $u_{t-1}$, e.g., (squared) Euclidean distance.  2) Temporal Variability: Defined as the variation in loss functions over time: $ V^f_T(\ell_{1:T}) \triangleq \sum_{t=2}^T \max_{a \in \mathcal{V}} | \ell_t(a) - \ell_{t-1}(a) | $ and $ V^g_T(\ell_{1:T}) \triangleq \sum_{t=2}^T \max_{a \in \mathcal{V}} \| \nabla\ell_t(a) - \nabla\ell_{t-1}(a) \|_2^2. $  Essentially, path length measures the cumulative change in the (optimal) decision over time, and temporal variability quantifies the maximum fluctuation in loss between consecutive time steps for any decision. A high path length indicates the optimal decisions vary significantly over time, while a high temporal variability reflects the extent to which the environment's feedback changes from one moment to the next, both of which signal a more volatile environment.\footnote{For simplicity, we do not discuss other performance measure variants (see \citep{rakhlin2011online} for an excellent exposition) or other nonstationarity measures.}


Online gradient descent (OGD) is arguably the most fundamental online learning algorithm. For introduction on this subject, we refer to \citep{shalev2012online,hazan2016introduction}; \citep{orabona2019modern} provides a unified view of many online learning algorithms as instantiations of Online Mirror Descent or Follow-The-Regularized-Leader and their variants 
\citep{orabona2019modern}. Existing approaches reveal strengths handling real-time data and certain environment changes \citep{hoi2021online}. However, open problems exist dealing with nonstationarity and distribution shifts, as discussed in Sec. \ref{sec:lowerbound}.


\section{Regret-Based Definition of Antifragility}
\label{sec:definition}

Antifragility, a concept going beyond resilience and robustness, refers to systems that don’t merely withstand disorder but benefit from it. 
This concept is particularly pertinent in the realm of online learning, where algorithms continually encounter shifting data streams and evolving decision contexts. Antifragility in this setting implies an algorithm’s ability not just to cope with these changes but to use them as a springboard for enhanced performance and adaptability.

Thus, in environments characterized by rapid and unpredictable changes, minimizing static regret is insufficient as it fails to account for the dynamic nature of the environment. Dynamic regret becomes vital as it compares system’s performance relative to the best possible action improves over time, despite frequent changes.


Let $U^K_T$ denote the set of comparator sequences $u_{1:T}$ where each comparator $u_t$ remains constant for every block of $K$ time steps and may change at the beginning of each subsequent block. Formally, 
\begin{align*}
    U^K_T = \big\{ &u_{1:T} \mid u_t =\tilde{u}_{(i)}, \; \quad\text{for} \; (i-1)K < t \leq iK, \; i \in[\mathrm{ceil}( {T}/{K})] \big\},
\end{align*}
where $\tilde{u}_{(i)}$ represents the comparator used for the $i$-th block of $K$ periods. For instance, if $K=2$, the sequences in $U^K_T$ will be in the form of $\tilde{u}_{(1)},\tilde{u}_{(1)},\tilde{u}_{(2)},\tilde{u}_{(2)},\ldots,\tilde{u}_{({T/2})},\tilde{u}_{({T/2})}$. In sequences from $U^K_T$, the volatility of the environment, as measured by path-length or temporal variability, is $\Omega(T/K)$, i.e., the system keeps changing. In this context, we provide a regret-based definition of antifragility of order $K$.

\begin{definition}[Order-$K$ Antifragility]
\label{def:antifragility}
A system is defined as \emph{order-$K$ antifragile} if it achieves sublinear dynamic regret for any sequence in $U^K_T$, i.e.,
\begin{equation}
    \label{def:antifragility}
    R_T^d(u_{1:T})=h(V_T), \quad \forall\;u_{1:T} \in U^K_T,
\end{equation}
where $K$ is the \emph{order} and $h$ is a non-negative \emph{strictly concave} function uniformly dominated by the identity function. We refer to $h$ as the variability-response function.
\end{definition}


\begin{remark}
    The set inclusion relationship $ U^K_T \subseteq U^{K'}_T $ for any $ K' \leq K $ indicates that a system antifragile at a lower order (smaller $ K $) inherently possesses antifragility at higher orders (larger $ K $). Consequently, achieving order-1 antifragility against $ U^1_T $ where the comparator changes every time step is the most challenging. This scenario is not just about mere adaptation; it transcends to a level of understanding the system's evolution, implying a capacity for perfect prediction. Order-1 antifragility suggests a system's ability to anticipate and respond to the utmost level of environmental volatility. Thus, a system that achieves antifragility at order 0 sets a high benchmark.
\end{remark}

\begin{remark}
    A  potentially confusing point in various personal discussions is that the mention of \emph{sublinear} dynamic regret in existing literature is often under the condition that the variation budget is bounded or sublinear in $T$. For instance, if $V_T=O(T)$, then existing bounds in the form of $O(V_T^\alpha T^{1-\alpha})$ would be $O(T)$, which does not satisfy the antifragility condition (since the corresponding variability response function is linear). A few exceptions exist. For instance, \citep{cheng2020online} can be viewed as a reduction to static regret using smoothness constraint on dynamics (see also Sec. \ref{sec:pomdp} for a general framework).
\end{remark}

\begin{remark}
    There are some special cases where even though the variation is linear under one notion of nonstationarity, the system may have sublinear dynamic regret by taking advantage of some ``best of different worlds'' bounds since the variation is constant in a different notion (see the example from \citep{jadbabaie2015online}). We exclude these cases because we consider these cases as approximately stationary environment.
\end{remark}

\subsection{Relation to \citep{taleb2013mathematical}}

\citep{taleb2013mathematical} quantifies antifragility in terms of structural properties of a stochastic payoff distribution---specifically, the sensitivity of positive tail integrals to volatility changes---formalizing a distribution-centric characterization (see also \citep{taleb2023working}). This definition focuses on a single random variable. Our proposed dynamic regret-based antifragility definition adopts a learning-theoretic perspective applicable to nonstationary decision-making systems. The two formalizations offer complementary lenses---probabilistic versus adaptive. Taleb's notion also distinguishes downside robustness from upside gain, imposing theoretical limits on excessive volatility that causes irrecoverable failure. Additional aspects therein could help refine our definition further.

Our regret-based definition examines adaptive risks under volatility, aligned with Taleb's perspective on distributional effects of uncertainty. Recall that well-known online-to-batch techniques relate regret bounds to expected excess risks $\mathbb{E}_{\ell_t \sim \mathcal{P}}[\frac{1}{T}\ell_t(a)]$ w.r.t. loss distribution $\mathcal{P}$ (e.g. \citep{cesa2004generalization}, \citet[Thm. 5.1]{shalev2012online}, and recently \citep{lugosi2023online}), as well as high probability bounds (e.g., \citep{rakhlin2017equivalence,orabona2023tight,van2023high}). See also recent extensions to non-IID batch settings such as stationary mixing processes (e.g. Markov chains) \citep{agarwal2012generalization} and nonstationary stochastic processes \citep{tao2020online}. These bound the path-dependent error $\mathbb{E}_{\ell_{T+s}}[\ell_{T+s}(a)|\{\ell_t\}_{t=1}^T]$ over horizon $T+s$ suitable for forecasting metrics \citep{kuznetsov2017generalization}.

One intriguing question is how to directly relate antifragility's dynamic regret to batch learning notions. To avoid trivial bounds (e.g., arising from the greedy optimization strategy), we have to exploit the strictly concave property of the variability-response function. One viable path is considering improper batch learning with latent contextual information, corresponding to an online predictor that understands latent sequence variables; this is akin to the extension of the batch settings from Markov Decision Processes (MDPs) \citep{kuznetsov2017generalization} to Partially Observable MDPs (POMDPs) (Sec. \ref{sec:pomdp}).

\subsection{Further discussion}

The most compelling theoretical aspect of antifragility lies in achieving  \emph{strictly concave response to variability}. Current views in online learning regards this as implausible per existing lower bounds (Sec. \ref{sec:lowerbound}).  However, observable antifragility in natural and engineered systems necessitates reevaluating underlying assumptions, especially concerning nonstationary environments. This reshifts focus onto theories and algorithms capable of adapting to and benefiting from real-world complexities. 

See App. \ref{app:measure} for a potential framework to quantify antifragility.



\section{Insights from Lower Bound Analysis}
\label{sec:lowerbound}

Given antifragility's assertion on strictly concave response to variability (Definition \ref{def:antifragility}), we tackle a fundamental criticism that it is theoretically unachievable due to existing lower bounds. Hence, we provide more discussions on lower bound analyses to reveal core obstacles---unpredictable shifts under adversarial constructions and information-theoretic limits posed by unstructured function classes. Rather than attempting a comprehensive review, our focus is on reevaluating assumptions to uncover potential inductive biases that could exceed established mathematical worst-case bounds.

Unpredictable and even adversarial shifts pose a fundamental obstacle for antifragility. \citep{besbes2015non} construct cost sequences with randomized, unpredictable optima ensures no policy can reliably predict shifting minima.  Leveraging the minimax static regret bound from \citep{abernethy2008optimal}, \citep{zhang2018adaptive} construct a piecewise constant comparator sequence with $V_T^u$ adversarially chosen segments to yield a $\Omega(\sqrt{T(1+V^u_T)})$ lower bound. The reduction from general sequences to piecewise constants with post-hoc adversarial minimization drives the result (c.f., \citet[Eq. 19]{zhang2018adaptive}).

Information-theoretic factors also constrain achievable regret. \citep{campolongo2021closer} use extreme value distributions exhibit a $\Omega(V_T^\gamma)$ lower bound on dynamic regret for any $\gamma\in(0,1)$ .  This aligns with results showing sublinear dynamic regret necessitates variation vanishing relative to time horizon \citet[Prop. 1]{besbes2015non}.

\citep{baby2021optimal} derive an information-theoretic lower bound $\Omega(V_T^{u\; 2/3} T^{1/3})$ matching their upper regret bound using wavelet-based function estimation theory for Triebel and Besov classes \citep{donoho1998minimax}. Such flexible function spaces with considerable spatial inhomogeneity can model spiky bounded variation comparators.   However, certain assumptions may be overly pessimistic:
\begin{itemize}
    \item Independent wavelet coefficients (as in Besov space) may overstate unpredictability. Real functions exhibit structured smoothness coupling among wavelet coefficients beneficial for forecasting.
    \item The minimax argument uses the least favorable priors concentrated on functions with isolated sparse spikes---the hardest to estimate adaptively. Additional restrictions such as smoothness, periodicity, or curvature can reduce the complexity of the problem.
\end{itemize}
While no algorithm exceeds constructed adversarial unpredictability or estimation limits mathematically, practical settings with appropriate inductive biases could plausibly break theoretical barriers. Even adversarial settings may have unmodeled weaknesses to exploit. For instance:
\begin{itemize}
    \item POMDP setting and its variants: Incorporating latent, unobserved states can provide predictability to account for apparent variability. Policies could potentially leverage these hidden modes to anticipate regime shifts (see Sec. \ref{sec:pomdp}).
    \item Smoothness, stability, or contraction constraints: Rather than entirely arbitrary loss functions, incorporating properties of the underlying dynamical systems, e.g., smoothness \citep{cheng2020online} and contraction \citep{lin2023online},  may expose exploitable patterns within variability. 
    \item Explicitly incorporating predictability: Directly modeling predictable regime changes could bypass limitations of adversarial assumptions (e.g., \citep{rakhlin2013online,hall2013dynamical}).
    \item Meta-learning settings: Providing policies with a few steps of gradient feedback after losses are revealed can reasonably measure adaptation (see Sec. \ref{sec:meta}).
\end{itemize}
However, care is needed so constraints do not exclude responding to black swan events. Smoothness and predictability may preclude  ``surprise'' events in the modeling stage. Rather than the raw input space, it could be more prudent to impose structure in latent spaces (such as wavelet space \citep{donoho1998minimax}, which naturally decomposes functions into global and local structures). Predictability may be lacking for heavy-tailed distributions, so the meta-learning settings that emphasize adaptation, improvision, and recovery can provide complementary benefits.

We believe online learning frameworks remain a core paradigm, but retaining these formulations with use-inspired augmentations can further antifragility goals. For instance, inherent structure may have already manifested in physical systems due to natural laws. As an analogy, certain quadratic constrained quadratic programming problems known to be NP-hard in general become polynomial-time solvable when the underlying graph structure or optimization has exploitable chordal sparsity \citep{lavaei2011zero} or benign landscape \citep{ge2016matrix}. Predictability may also be addressed through changes outside purely algorithmic scope as part of the broader decision stack, such as increasing sensor observation frequency in hardware components. We advocate expanding modeling to system-level factors whenever suitable, whether stemming from problem geometry, physics, or hardware capacities.

\section{Significance and Impact}

Many biological organisms exhibit \emph{intrinsic} antifragility, gaining strength from stressors such as hypoxic conditioning or bioenergetic regeneration \citep{taleb2023working}. Incorporating antifragility principles into clinical practice can boost effectiveness in areas such as cancer treatment,  nutrition cycles, and exercise regimens exercise regimens (see \citet[Table A1]{taleb2023working}).

However, sociotechnical infrastructure still lacks flexible adaptation to mounting volatility and looming black swans. Modern power systems face increasing supply and demand uncertainties, amplifying extreme stress events risks. But legacy robustness cannot manage cascading outage threats. Despite smart grid enhancements, fragility persists with 60\% of assessment criteria indicating high likelihood of performance degradation \citep{johnson2013antifragility}. Antifragility can potentially be used to improve cybersecurity, e.g., to detect zero-day attacks \citep{manzoor2024zda}, or power system resilience by improving fast response rate to critical load restoration \citep{abdeen2024clrmeta}.

Similar reliability challenges under long tails emerge in areas like epidemic control, supply chains, and agricultural ecosystems. Even human collectives struggle with rare yet impactful events, as the 2008 crisis and COVID-19 pandemic. This blueprint is to pioneer ML tools to computationally induce antifragility where intrinsically lacking.

\section{Towards Computational Antifragility
}
\label{sec:discussion}

This section identifies ML subareas that hold potential for cross-pollination of ideas towards engineering antifragility in ML systems. The goal is sparking discussion by exemplifying useful mechanisms and surfacing gaps through a non-comprehensive but targeted sampling of domains.

\subsection{Nonstationary online learning: Forecasting shifts}
\label{sec:nonstation}
Nonstationary online learning is a foundational setup for studying antifragility. As discussed in Sec. \ref{sec:lowerbound}, reactive mechanisms (e.g., sliding window) without consolidating experience in generalizable knowledge (e.g., restarting mechanism) may face fundamental limits. 

A promising direction is to adopt a proactive approach by synergizing with the field of forecasting, which has abundant important results dealing with nonstationarity (e.g., \citep{kuznetsov2015learning,ansari2023neural}). Prior works such as \citep{rakhlin2013online} provide a viable reduction approach that can potentially achieve sublinear dynamic regret if forecasting models demonstrate antifragility. For nonstationary settings, recent works explore forecasting value functions \citep{chandak2020optimizing}, perturbations \citep{yu2020power}, and models \citep{lee2023tempo}, showing promising empirical/theoretical results. Another direction is multimodal approach (see \citep{liang2023tutorial} for a recent survey), as certain modes (e.g., language) may exhibit more stationarity or support large-scale diversified data for better forecasting.

\subsection{POMDP and variants: Modeling and identifying nonstationarity}
\label{sec:pomdp}
POMDPs provide a mathematical framework for sequential decision making under uncertainty and partial observability with diverse real-world applications \citep{kurniawati2022partially}. POMDPs can capture nonstationarity as hidden state evolution, epistemic uncertainty as state ambiguity, and aleatoric uncertainty as inherent stochastic transitions/rewards. The potential reduction of dynamic regret to a static regret with respect to an optimal policy that plans with oracle access to hidden state means algorithms in this setting may potentially handle order-$K$ antifragility (e.g., $K$ is the interactions needed for inferring latent state as in decodable POMDP \citep{efroni2022provable}).

POMDPs are NP-hard in general \citep{papadimitriou1987complexity,vlassis2012computational}, since they involve reasoning about all possible belief states of the environment. Thus, recent works have focused on more structured, simplified yet realistic settings.  The Latent MDP (LMDP) \citep{kwon2021rl} assumes an unrevealed latent factor identifying the MDP, closely relating to predictive state representations (PSR) \citep{singh2004predictive} that directly model future observation probabilities conditional on past observations.
\citep{lee2023learning} considers a similar setting called Hindsight Observable Markov Decision Process (HOMDP) where the latent states are revealed to the learner in hindsight and provides regret bounds $O(X)$ where $X$ is the number of latent states. Block MDPs \citep{krishnamurthy2016pac,du2019provably} and decodable POMDP \citep{efroni2022provable} are special classes of POMDPs that enable current (or recent) observations to exactly decode current latent states with some hidden decoding function. Other related formulations include hidden-model MDPs \citep{chades2012momdps}, multitask RL \citep{brunskill2013sample}, multi-model MDPs \citep{steimle2021multi}, contextual MDPs \citep{hallak2015contextual}, factored MDPs \citep{huang2021adarl,feng2022factored}, and Bayes-adaptive MDPs \citep{zintgraf2019varibad}; strategies include clustering \citep{brunskill2013sample,kwon2021rl}, variational inference \citep{zintgraf2019varibad,yang2019single,ghavamzadeh2015bayesian, huang2021adarl}, and causal inference \citep{feng2022factored}.

A key insight is the necessity of memory-based policies that leverage histories to reason about latent states at test time. Existing methods advance towards antifragility in the sense that if the number of latent states is finite or sublinear, then the dynamic regret (when viewd as the static regret with respect to optimal POMDP policy) is sublinear regardless of switching variation budgets \citep{kwon2021rl,efroni2022provable}; however, they fall short in lifelong learning under continuous perturbation that introduces new states. Additionally, while some test-time adaptation is enabled through training diversity, assumptions of a closed-world post-training limit adaptation capabilities. While some broader settings have been examined (c.f., \citep{zhong2022posterior} that include POMDPs and PSRs and provide conditions for sample efficient learning using generic posterior sampling algorithms), identifying realistic conditions motivated by use cases that still allow computationally and statistically tractable learning for antifragility remains an open question. Finally, causal RL that identifies the latent change factors represents a promising direction for fast adaptation and interpretability \citep{zeng2023survey}.



\subsection{Safe/Robust learning: Expanding boundaries}
\label{sec: safe}
Safe learning aims to provide guarnatees that policies uphold constraints on avoiding unsafe regions or performance thresholds \citep{gu2022review,brunke2022safe}. Antifragility necessitates cautiously expanding safety boundaries, which can be achieved via: 
\begin{itemize}
    \item Safe exploration that predicts safety risks \citep{thomas2021safe,wachi2023safe}, under guarantees of stability \citep{berkenkamp2017safe} or returnability \citep{eysenbach2018leave}.
    \item Transferrable safety, like safety critics \citep{srinivasan2020learning,bharadhwaj2020conservative} or safe meta-policy adaptation \citep{khattar2022cmdp}, to substantially reduce incidents through past experience.
    \item Learning-based control-theoretic methods such as reachability analysis for returnability \citep{bansal2017hamilton}, robust control such as integral-quadratic constraints \citep{megretski1997system,gu2022recurrent}, model-predictive control \citep{hewing2020learning,brunke2022safe}, control-barrier functions \citep{ames2019control,dawson2023safe}, and neural Lyapunov methods \citep{dawson2023safe}.
\end{itemize}

\subsection{Continual learning: Balancing plasticity \& stability}
\label{sec:continual}
Continual learning (also called incremental/endless/lifelong learning and closely related to nonstationary learning) refers to the ability of acquiring, updating, accumulating, and exploiting knowledge throughout a system's lifetime \citep{lesort2020continual,de2021continual,khetarpal2022towards}. The key challenge is catastrophic forgetting---adapting to new distributions reduces prior capabilities. \citep{kim2022theoretical} decomposes continual learning into within-task prediction and task-id prediction (related to OOD detection) as necessary and sufficient conditions, providing a constructive reduction approach. Though foundation models may suffer less forgetting \citep{ramasesh2021effect}, antifragility benefits from techniques balancing plasticity and stability (e.g., via regularization and experience replay), especially when updating models as novelty inevitably arises. Also, the field has recently focused on generalization \citep{wang2023comprehensive}, a crucial element for antifragility to enable improvisation.

\subsection{Meta learning: Fast adaptation}
\label{sec:meta}
Meta-learning (also called learning-to-learn) focuses on rapidly adapting to new tasks by acquiring reusable knowledge about the learning process itself \citep{hospedales2021meta,beck2023survey}. Approaches can be generally viewed as a primary (outer level) system progressively improving the learning of a secondary (inner level) system, where the inner system performs sequentially arriving tasks while the outer system does adaptation using algorithms like RL \citep{duan2016rl,wang2016learning}, online learning \citep{khodak2019adaptive}, Blackwell approachability strategies \citep{niazadeh2021online}, and evolution strategies \citep{houthooft2018evolved,lu2022discovered,khattar2023winning}. Memory-based meta-learning offers computational shortcuts to difficult probabilistic inferences for acceleration \citep{ortega2019meta} (c.f., Sec. \ref{sec:pomdp}). In-context learning (ICL) \citep{brown2020language} represents a recent paradigm that does not require parameter updates at test time (see Sec. \ref{sec:pretraining}).

Black swan events often involve solving hard optimization problems due to the high-dimensional search space and nonconvex landscape; learning a representation such as a latent low-dimensional manifold \citep{sel2023learning} or latent variable identifier \citep{yang2019single} can facilitate efficient search and even single attempt success. Furthermore, while recent works such as \citep{khattar2022cmdp} exhibit certain properties of $K$-order antifragility by embedding a sequence of safe RL problems within an online learning framework (where $K$ is the number of within-task steps) in an adversarial setting, further considering the predictability of the environment and leveraging memory in the primary learning system can help break the lower bound for the reason outlined in Sec. \ref{sec:lowerbound}. An important goal is to perform well in a single attempt,

\subsection{Foundation models: Pretraining and memory}
\label{sec:pretraining}

Foundation models have shown impressive capabilities in various problems \citep{bommasani2021opportunities}, which crucially depend on pretraining on vast and diverse tasks to gain on-the-fly adaptation, i.e., in-context learning (ICL). Recent works extend this paradigm to online decision making such as adaptive agent (AdA) \citep{bauer2023human}, algorithm distillation \citep{laskin2022context}, Decision-Pretrained Transformer (DPT) \citep{lee2023supervised}, and Gato \citep{reed2022generalist}. Notably, the training resembles next-token prediction and is often as simple as supervised learning with autoregressive loss \citep{laskin2022context,reed2022generalist,lee2023supervised} (though the history-dependent policy can be viewed as a computational posterior sampling for POMDPs \citep{osband2013more,lee2023supervised}).

For test-time adaptation, memory architectures such as Neural Turing Machines \citep{graves2014neural} offer a direct approach, with learning involving discovery and encoding of abstract rules. Retrieval-augmented generation (as surveyed in \citep{gao2023retrieval,zhao2023retrieving}) offers prospects for rapidly selecting and improvising from existing skill repertoires.

\subsection{QD and MO learning: Adaptive repertoire}
\label{sec:qdmo}

A key imperative for managing OOD events is rapidly searching for compensatory behaviors in unfamiliar scenarios, such as a robot adopting alternatives when damaged \citep{cully2015robots}. Antifragile systems thus have to judiciously leverage natural perturbations to gather diverse response repertoires.

Quality-diversity (QD) formalizes such expansion---optimizing solution archives over behavior descriptors to maintain diversity \citep{chatzilygeroudis2021quality}. Mathematically, given a solution space $\Pi$, QD considers an objective $f_0(\pi)$ for $\pi\in \Pi$ and behavior descriptors $f_i(\pi)$, $\forall i\in[k]$ capturing solution properties, e.g., reversibility, information gathering, robustness, safety compliance, and simplicity. Recent methods move towards more open-ended domains with automatic descriptor discovery using low-dimensional embeddings \citep{grillotti2022unsupervised}, with feedback provided by human \citep{ding2023quality} or AI \citep{bradley2023quality}. For antifragility, online QD by synergizing with continual/meta/online learnings can dynamically maintain a diverse repertoire to retrieve or improvise appropriate strategies, e.g.,  substituting alternative behaviors when physical capacities degrade; switching to safety-preserving actions on constraint changes; activating uncertainty-robust policies amid chaos; and recruiting streamlined approaches when data is scarce post-disruptions.

Multiobjective (MO) learning provides a set of tools that can complement QD to balance potentially conflicting behavioral descriptors \citep{yu2020gradient,liu2021conflict,gu2024balance}, directly align objectives with preferences \citep{mahapatra2020multi,chen2022multi}, and maintain a set of Pareto optimal policy repertoire for test-time adaptation \citep{yang2019generalized,navon2020learning,lin2022pareto,kyriakis2022pareto,dimitriadis2023pareto}.

\subsection{Adversarial ML: Vaccination by attack}
\label{sec:adversarial}

Adversarial ML involves exposing systems to deliberately designed inputs to induce failures and understand robustness/fragility \citep{hendrycks2021unsolved}. In addition to existing robustness benchmarks \citep{hendrycks2019benchmarking,hendrycks2021many,koh2021wilds}, researchers could create more environments to stress-test systems (e.g., autonomous driving \citep{tang2023survey}) and estimate the antifragility regret. For instance, competitions like the AI Village at DEFCON conferences enable wide discovery of harms from public participation to address them \citep{whitehouse24red}. Using such benchmarks and environments, researchers could improve ML systems to withstand black swans, long tails, and structurally novel events. Another interesting direction could be antifragility training, where exposing systems to a variety of attacks during training may lead to a better understanding of potential new attacks, as explored by \citet{manzoor2024zda} for zero-day attack detection.

\section{Call for Actions and Future Directions}
\label{sec:future}

To operate reliably in open-world high-stakes environments, ML systems need to endure unforeseen events and tail risks. While there has been significant progress in ODM under nonstationarity, vast opportunities remain untapped for channeling volatility into intrinsic asset rather than solely a fragility risk.  This necessitates an open-system perspective that sustains safety expansions under distributional shifts. Antifragility constitutes pioneering conceptual and computational upgrades to mitigate long-term risks, focusing on \emph{unusual robustness} against black swan events.

Significant open challenges remain in architecting decision systems to reliably \emph{improve} under distributional shifts. Core future pursuits involve:
\begin{itemize}
    \item Developing practical algorithms actualizing key antifragility principles, such as leveraging short-term predictability and embracing small, controlled failures/risks hedge against catastrophic failures, accumulating options and strategic flexibility, rapidly adapting and agile improvision, distilling invariant structures and generalizable patterns from diverse experience, and proactive inoculation against outliers with an emergence of generalization capabilities.
    
    \item Formalizing foundations for antifragility guarantees grounded in learning theories so systems demonstrably gain from volatility through principled mechanisms, while ensuring safe exploration and adaptation. This requires characterizing the ``value of volatility'' (c.f., the psychology counterpart of challenge-hindrance framework \citep{o2019so}) for selective exposures under controlled experimentations.
    
    \item Architecting entire decision-making stack (spanning ML components, physical computing infrastructure, and downstream applications) with end-to-end antifragility. Priorities involve composing modular system architectures that sustain diversity of observational sensing, behavioral responses, and safety constraints. This prevents isolated fragilities from cascading across layers and retains operation despite disruptions. Foundation models have potential for transfer and adaptation but require expanded competencies and consolidation mechanisms to mitigate negative interference and catastrophic forgetting.

    \item Developing  benchmarks and simulation environments that calibrate controllable uncertainty injections for reliably quantifying volatility-related performance improvements, especially as measured by long-tail high-impact events.
\end{itemize}
Progress demands constructive engagement across computational subfields of ML, optimization, and control communities, deliberate involvement of interdisciplinary researchers, and proactive participation with stakeholders.

\bibliography{example_paper}
\bibliographystyle{icml2024}

\appendix
\section{Concrete examples and empirical evidence}
\label{app:examples}

\emph{Biological systems:} Evolution is inherently antifragile---pressures prompt adaptations enabling greater species resilience over generations, like bacteria developing stronger resistance to antibiotics designed to eliminate them. Tropism in plants allow dynamically bending towards beneficial stimuli like light, enhancing robustness despite variability. Even mild climate changes may elicit adaptive responses in coral resilience \citep{hughes2003climate}.  Ecosystems with higher biodiversity demonstrate greater adaptability to environmental changes \citep{folke2004regime}, and can be actively leveraged to improve resilience, e.g., prescribed fire \citep{ryan2013prescribed}. The immune system strengthens from exposure to pathogens (e.g., through vaccination \citep{plotkin2005vaccines}), and skeletal muscles grow in response to the moderate stress of exercise \citep{schoenfeld2010mechanisms}.

\emph{Economic systems:}  Decentralized markets, characterized by price fluctuations and competition, drive innovation and efficiency \citep{hayek2013use}. Entrepreneurship often benefits from failure and adversity, leading to future success \citep{mcgrath1999falling}. Some investment strategies, such as antifragile portfolios like ``long vega'' and ``long gamma'' financial derivatives, are designed to profit from market volatility \citep{taleb2013mathematical}.

\emph{Social systems:} Collective intelligence, which relies on the diversity of opinions and experiences (viewed as internal opinion ``stress testing''), enhances problem-solving and decision-making capabilities \citep{page2008difference}. Resilient communities adapt and thrive under challenge \citep{norris2008community}.

\emph{Technological systems:} Agile development allows rapid response to changing requirements \citep{manifesto2001manifesto}.

\emph{Psychology:} Adversity can lead to post-traumatic growth \citep{tedeschi2004posttraumatic}. Hormesis \citep{mattson2008hormesis}, where low doses of toxins trigger beneficial responses, is another example.

\emph{Engineering systems.} Early steam engines advanced from fragile explosiveness to reliable operations due to an engineering discovery---intentionally introducing randomness (dithering) stabilized operations by overcoming mechanical stiction. Chaos Control theory explores how the principles of chaos theory can be applied to engineering systems to achieve faster control and stability \citep{ott1990controlling}. The concept of using noise for stabilization in early steam engine design, known as ``stochastic resonance'' or ``noise-induced stability,'' is related to the principles of Chaos Control theory \citep{gammaitoni1998stochastic}. By understanding and leveraging the properties of chaotic systems, engineers can design more efficient and responsive control systems that are agile, adaptive, and capable of rapidly responding to changes in their environment \citep{scholl2008handbook}.

While some sources may use terms like ``resilience,'' these examples go beyond mere recovery. Under stress, a resilient system bounces back; an antifragile system bounces forward, returning to a state stronger than before. This distinction highlights the potential benefits of designing with antifragility in mind.

\section{FAQs}

\subsection{How to quantify and measure antifragility? }
\label{app:measure}
While antifragility can be rigorously checked in theoretical settings (based on the proposed antifragility definition), assessing it in practice requires a comprehensive approach. To establish antifragility as a system property, it is insufficient to observe the system under a single trajectory. Instead, we need to systematically expose the system to different levels of stress (controlled experiments) and observe its performance across various settings and case studies
\begin{itemize}
    \item Use high-fidelity simulations when real-world experiments are limited.
    \item Analyze system responses to past stress in settings like hindsight POMDPs or continual learning with immediate feedback. Define stress as diversity and coverage of high-impact novel examples. There are also existing works in medical research and quantifative finance by \citep{taleb2013mathematical,taleb2018anti,taleb2023working} and co-authors that attempt to quantify antifragility using various case studies.
    \item Collaborate to define benchmarks, design experiments, and build simulations.
    \item Leverage ML to learn variability-performance relationships or identify patterns under different stress levels, uncovering concave responses from historical or simulated data.
\end{itemize}

\subsection{What are the trade-offs in applying antifragility for decision systems?
}

Antifragility is a long-term property focusing on thriving after experiencing stressful conditions, while efficiency, interpretability, and fairness are often considered under normal conditions. These properties focus on different system aspects:
\begin{itemize}
    \item Antifragile systems may trade efficiency for redundancy, diversity, and adaptability. However, antifragility can complement normal-condition efficiency with ``context-dependent response mechanisms'' for fast mode switching, resulting in less pronounced trade-offs. Selectively applying antifragility to critical parts can help.
    \item Complex antifragile mechanisms can be difficult to explain, but interpretability can be potentially incorporated by design using XAI, rule-based systems, or human-in-the-loop approaches.
    \item Antifragile systems learning under stress might inadvertently amplify biases or discriminatory patterns. Techniques to mitigate bias, ensure data diversity, maintain balanced risks across regions, and regular audits are important.
\end{itemize}

\subsection{How is antifragility related to paradigms such as robustness, lifelong learning, and cognitive science?}
\label{app:related}
Antifragility emphasizes \emph{unusual} robustness, with systems proactively seeking out and leveraging (virtual) stressors for open-ended learning and continuous improvement. This approach enables a potentially more effective response to perturbations outside the typical range of expected scenarios (i.e., black swans), which may correspond to unknown attacks in adversarial ML. Designing antifragile learning strategies requires understanding the nature of stressors, system dynamics, and desired outcomes. It may also hint at a solution for breaking the defense-attack arms race by potentially enabling the emergence of defense capabilities through antifragility training (e.g., \cite{manzoor2024zda}).

Lifelong and open-world learning approaches share similarities with antifragility but can benefit from incorporating its principles to enhance their ability to handle unexpected events and benefit from stressors. Antifragile systems, such as the immune system (through vaccination \citep{plotkin2005vaccines}) and ecosystems (through prescribed burns \citep{ryan2013prescribed}), demonstrate the value of controlled perturbations for increased resilience.

Antifragility focuses on decision-making scenarios where the magnitude of wrong decisions is important and on unusual robustness against extreme perturbations, an aspect not fully addressed in existing robustness or open-ended/continual learning.

Cognitive science paradigms like counterfactual simulation \citep{roese2017functional} share common ground with antifragility, but the latter puts these ideas on a more theoretical foundation and addresses the computational aspects of designing systems that benefit from uncertainty.

\section{Ethical Considerations}
\label{app:safe}

\subsection{Addressing potential harms of antifragility training for LLMs}

When applying antifragility principles to the training of ML models, such as large language models (LLMs), it is crucial to prioritize ethical considerations to prevent harm and ensure socially responsible development. Key ethical considerations include:
\begin{itemize}
    \item Ensuring fair treatment, appropriate compensation, and protection of human workers involved in providing feedback or data for model training.
    \item Carefully curating and vetting training data and inputs to avoid exposing models or human workers to harmful, offensive, or biased content.
    \item Maintaining transparency in model development processes, regularly auditing models for potential biases or harmful outputs, and establishing clear accountability measures.
    \item Engaging with diverse stakeholders, including ethicists, policymakers, and community representatives, to identify potential risks, unintended consequences, and align model development with societal values.
    \item Continuously monitoring model performance and impacts on users and society, promptly addressing any signs of harmful or biased outputs, and refining the model and training processes accordingly.
\end{itemize}

\subsection{Challenges and mitigation strategies for safety-critical domains}

Antifragile approaches may introduce unacceptable risks in safety-critical domains. Key challenges include:
\begin{itemize}
    \item Strict regulations and standards that may be violated by intentional stress exposure.
    \item Difficulty determining safe stress levels. Too much stress is risky; too little may not provide sufficient opportunities for improvement.
    \item Controlled experiments key to antifragility may be impossible due to safety concerns.
    \item Safety-critical systems are interconnected, with multiple subsystems interacting in intricate ways. Introducing antifragile mechanisms in one part may have unintended consequences or cascading effects on other parts.
    \item Stakeholder resistance to intentionally stressing systems.
\end{itemize}

\noindent To harness antifragility benefits while ensuring safety:
\begin{itemize}
    \item Leverage simulations and digital twins for safe exploration.
    \item Thoroughly assess risks and have strong mitigation plans. This includes identifying potential failure modes, defining safety boundaries, and establishing contingency plans to prevent or limit the consequences of failures.
    \item Introduce antifragility gradually, starting with low-risk parts and monitoring closely.
    \item Capitalize on inherent stress from historical failures and natural disruptions, using causal reasoning to maximize data value.
    \item Continually adapt and adjust mechanisms based on feedback.
    \item Last but not least, build trust with stakeholders by clearly explaining how the system adapts. This requires close collaboration among researchers, practitioners, regulators, and stakeholders to ensure that antifragile systems are designed and implemented in a responsible and context-appropriate manner.
\end{itemize}

\end{document}